\documentclass[10pt,journal,compsoc]{IEEEtran}
\ifCLASSOPTIONcompsoc
\usepackage[nocompress]{cite}
\else
\usepackage{cite}
\fi
\ifCLASSINFOpdf
\usepackage[pdftex]{graphicx}
\else
\fi

\usepackage{amsmath}
\usepackage{algorithm}
\usepackage{algorithmic}
\usepackage{array}
\ifCLASSOPTIONcompsoc
\usepackage[caption=false,font=footnotesize,labelfont=sf,textfont=sf]{subfig}
\else
\usepackage[caption=false,font=footnotesize]{subfig}
\fi
\usepackage{stfloats}
\usepackage{url}

\usepackage{float}
\usepackage{cite}
\usepackage{amssymb}
\usepackage{slashbox}
\usepackage{epsfig}
\usepackage{graphicx}
\usepackage{babel}
\usepackage{threeparttable}
\usepackage{booktabs}
\usepackage{makecell}
\usepackage{multirow}
\usepackage{color}
\usepackage{float}
\usepackage{stfloats}
\usepackage[colorlinks,linkcolor=blue,anchorcolor=blue,citecolor=blue]{hyperref}
\usepackage{algorithm}
\usepackage{algorithmic}
\usepackage{footnote}
\usepackage{diagbox}

\newcommand{\tabincell}[2]{\begin{tabular}{@{}#1@{}}#2\end{tabular}}

\begin{document}
\title{Pair-Relationship Modeling for Latent Fingerprint Recognition}
\author{
        Yanming~Zhu,
        Xuefei~Yin,
        Xiuping~Jia,~\IEEEmembership{Fellow,~IEEE}
        and~Jiankun~Hu$^{*}$,~\IEEEmembership{Senior Member,~IEEE}
\IEEEcompsocitemizethanks{\IEEEcompsocthanksitem Yanming Zhu is with School of Computer Science and Engineering, University of New South Wales, Sydney, NSW 2052, Australia (e-mail: yanming.zhu@unsw.edu.au).
\IEEEcompsocthanksitem Xuefei Yin, Xiuping Jia, and Jiankun Hu are with School of Engineering and Information Technology, University of New South Wales, Canberra, ACT 2600, Australia (e-mail: xuefei.yin@unsw.edu.au; x.jia@adfa.edu.au; $^*$Corresponding author: J.Hu@adfa.edu.au).}
%\thanks{Manuscript received April 19, 2005; revised August 26, 2015.}
}
% The paper headers
%\markboth{Journal of \LaTeX\ Class Files,~Vol.~14, No.~8, August~2015}%
%{Shell \MakeLowercase{\textit{et al.}}: Bare Demo of IEEEtran.cls for Computer Society Journals}

\IEEEtitleabstractindextext{
\begin{abstract}
Latent fingerprints are important for identifying criminal suspects. However, recognizing a latent fingerprint in a collection of reference fingerprints remains a challenge. Most, if not all, of existing methods would extract representation features of each fingerprint independently and then compare the similarity of these representation features for recognition in a different process. Without the supervision of similarity for the feature extraction process, the extracted representation features are hard to optimally reflect the similarity of the two compared fingerprints which is the base for matching decision making. In this paper, we propose a new scheme that can model the pair-relationship of two fingerprints directly as the similarity feature for recognition. The pair-relationship is modeled by a hybrid deep network which can handle the difficulties of random sizes and corrupted areas of latent fingerprints. Experimental results on two databases show that the proposed method outperforms the state of the art.
\end{abstract}

}

\maketitle
\IEEEdisplaynontitleabstractindextext
\IEEEpeerreviewmaketitle

\begin{figure}[tbp]
	\centering
	\includegraphics[width=0.8\linewidth]{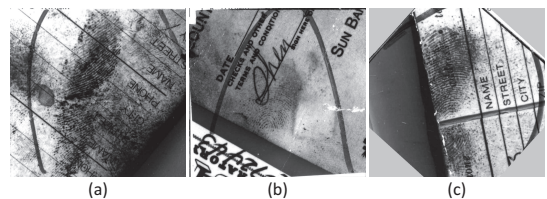}
	\vspace{-0.12in}
	\caption{Samples of latent fingerprints obtained from the NIST SD27 database: (a) good quality; (b) bad quality; and (c) ugly quality.}
	\label{fig_no1}
	\vspace{-0.15in}
\end{figure}

\IEEEraisesectionheading{\section{Introduction}}
\IEEEPARstart{B}{iometric} authentication has been used prominently in human verification and identification. Of the many biometrics, fingerprint has proven to be a very reliable index, and has been widely used as evidence by law enforcement and forensic agencies \cite{2MaltoniHandbook}. The three types of fingerprints in law enforcement applications are rolled, plain, and latent. The first two are obtained intentionally by rolling or placing a finger on a piece of paper or scanner platen while the last are lifted from finger impressions unintentionally left by a person at a crime scene. Compared with plain and rolled fingerprints, a latent one is usually smudgy and blurred, showing only a small finger area and usually a large non-linear distortion. As shown in Fig. \ref{fig_no1}, these fingerprints are of low quality, with unclear ridge structures, complex background noise, and various overlapping patterns. Therefore, the performance of a fingerprint identification system for latent fingerprints is much poorer than that for rolled or plain ones \cite{3WatsonFingerprint,4IndovinaEvaluation}. 

In the past decades, many efforts have been made towards recognizing latent fingerprints. The developed methods can be basically classified into two categories: 1) traditional methods and 2) learning-based methods. Using more manual features has been widely reported improving recognition accuracy in early traditional methods \cite{8jain2011latent,9feng2009latent,10jain2008matching,11001arora2014latent}. However, manually marking features is very time-consuming/labor-intensive and possibly infeasible for low-quality fingerprints. Using a reduced number of manual features but constructing more complex strategies or more efficient algorithms is another choice in traditional methods \cite{17yoon2011latent,18paulino2013latent,19001medina2016latent}. 
However, these techniques face significant challenges because only a small proportion of latent fingerprints are available for their feature extraction. With the great success of machine learning for various recognition tasks, learning-based methods have been proposed for latent fingerprint recognition \cite{nguyen2018robust,21cao2018automated,cao2018latent,cao2018end}, which aim to extract more robust features to represent fingerprints to improve the recognition accuracy. Generally, these methods still follow the traditional way of extracting representation features for each fingerprint independently and then compare the similarity of these representation features for recognition in a different process. Thus, useful correlations of the two compared fingerprints are lost during the representation feature extraction process. Once they are lost, they cannot be recovered in the recognition process. Also, as the representation feature extraction process and the recognition process are separate and independent, they cannot be jointly optimized. Without the supervision of similarity, the extracted representation features are hard to optimally reflect the similarity of the two compared fingerprints which is the base for matching decision making. 

Therefore, in this paper, we propose a new scheme for latent fingerprint recognition. Instead of extracting representation features of each fingerprint independently and then comparing the similarity of these representation features for recognition in a different process, we propose to directly model the pair-relationship\footnote{The pair-relationship represents the similarity of two fingerprints (/fingerprint patches).} of two fingerprints as the similarity feature for recognition. Hence, correlations of two fingerprints can be exploited for better characterization of the similarity which is the base for matching decision making. Specifically, the pair-relationship is modeled by a hybrid deep network using tensors of two fingerprints which are concatenations of two fingerprint images and concatenations of their orientation fields. The hybrid deep network consists of a set of convolutional neural networks (CNNs) and a restricted Boltzmann machine (RBM). The CNNs are designed to have different sizes and architectures to handle the difficulties of random sizes and corrupted areas of latent fingerprints. The RBM is designed to make a decision based on the complementary outputs of the CNNs. Experimental results on two databases show that the proposed method outperforms the state of the art.

The main contributions of this paper are as follows.
\begin{itemize}
	\item[1)] A new idea is proposed for latent fingerprint recognition. Unlike the current methods that extract representation features of each fingerprint independently and then compare the similarity of these representation features for the recognition in a different process, we directly model the pair-relationship of the two fingerprints as the similarity feature for the recognition.

    \item[2)] A hybrid deep network is proposed to model the pair-relationship, which is composed of a set of CNNs and an RBM. The set of CNNs is delicately designed to have different sizes and architectures and thus can model the pair-relationship from multiple information levels and handle the difficulties of random sizes and corrupted areas of latent fingerprints. The matching decision is made by the RBM based on the complementary outputs of these CNNs, which is more effective.
    
	\item[3)] Most of the existing learning-based methods extract representation features directly from fingerprint images. The proposed hybrid deep network extracts complementary pair-relationship from fingerprint images and their orientation fields regularized by the FOMFE model \cite{wang2007fingerprint,RN759} which provides a function of denoising and estimating orientation fields for moderately corrupted areas. 
	
	\item[4)] A new concept of macro-patch is introduced. Together with the innovative latent-rolled fingerprint tensor, the pair-relationship can be effectively encoded in the presence of distortion and corrupted areas in the latent fingerprints. 

	\item[5)] A new M-DLO alignment is proposed based on the DLO alignment \cite{yager2005coarse}, which provides more accurate alignment in the challenging latent fingerprint environment. 
\end{itemize}

The rest of this paper is organized as follows: Section \ref{sec:relatedworks} reviews related works on latent fingerprint recognition; Section \ref{sec:proposedmethod} details the proposed method; experimental results and discussions are presented in Section \ref{sec:experimentalresults} and, finally, the paper is concluded in Section \ref{sec:conclusion}.

\vspace{-0.07in}
\section{Related Works}
\label{sec:relatedworks}
Over the last few decades, many methods for dealing with the inherent complexity of latent fingerprints and improving recognition accuracy have been proposed, as surveyed in \cite{22001sankaran2014latent}. Early methods used many manually marked features to improve recognition accuracy. For example, Jain et al. \cite{10jain2008matching} used minutiae, orientations, and quality maps for the recognition. Feng et al. \cite{11feng2008filtering} used minutiae, pattern types, orientation fields, reference points, and singular points for the recognition. Jain et al. \cite{8jain2011latent} used seven types of features to improve recognition accuracy. Although these methods achieve good recognition performance, they are laborious and time-consuming due to their manual feature marking. Also, the repeatability of this form of feature markup is low and thus limits the practical application of these methods.

Later, methods of using a reduced number of manual features have been proposed. For example, Yoon et al. \cite{17yoon2011latent} proposed a method which requires only manually marked ROI and singular points for the recognition. Paulino et al. \cite{18paulino2013latent} proposed a method based on a descriptor-based Hough transform for the recognition, where only minutiae are manually marked. Medina-Perez et al. \cite{19001medina2016latent} proposed a method based on clustering, which aims to improve its comparison algorithm while not adding extended features, achieves impressive results in the presence of large non-linear deformations in latent fingerprints. In general, these methods attempt to construct more complex structures of features or seek more efficient comparison strategies for latent fingerprint recognition. Their performance heavily depend on the accuracy of the few features they used, and extracting these features is difficult and even infeasible for many latent fingerprints.

Recently, with the superiority of deep learning for various recognition tasks, learning-based methods have been proposed for latent fingerprint recognition. For example, Cao et al. \cite{21cao2018automated} proposed using CNNs to estimate ridge flows and extract minutiae descriptors, and then extracting minutiae and texture templates to represent fingerprints for the recognition. Cao et al. \cite{cao2018latent} further improved the template extraction in \cite{21cao2018automated} to better represent fingerprints for the recognition. Cao et al. \cite{cao2018end} proposed using autoencoders to enhance a latent fingerprint and extract its minutiae, and then using CNNs to extract minutia descriptor and texture template to represent fingerprints for the recognition. In general, these methods follow the traditional way of extracting representation features for each fingerprint independently and then comparing the similarity of these representation features in a different process. The representation features used to represent the fingerprint are learned without the supervision of similarity. These methods achieve impressive results, yet the performance can be further improved.

Therefore, in this paper, we propose a new scheme that can directly model the pair-relationship of two fingerprints as the similarity feature for recognition. Consequently, correlations of two fingerprints are exploited for better characterization of the similarity which is the base for matching decision making. The similarity feature is learned with the supervision of recognition which improves its robustness. Experimental results on two databases show that the proposed method outperforms the state of the art.

\vspace{-0.1in}
\section{Proposed Method}
\label{sec:proposedmethod}
\subsection{Overview of Framework}
\label{subsec:frameworkoverview}
\begin{figure*}[htbp]
	\centering
	\includegraphics[width=0.78\linewidth]{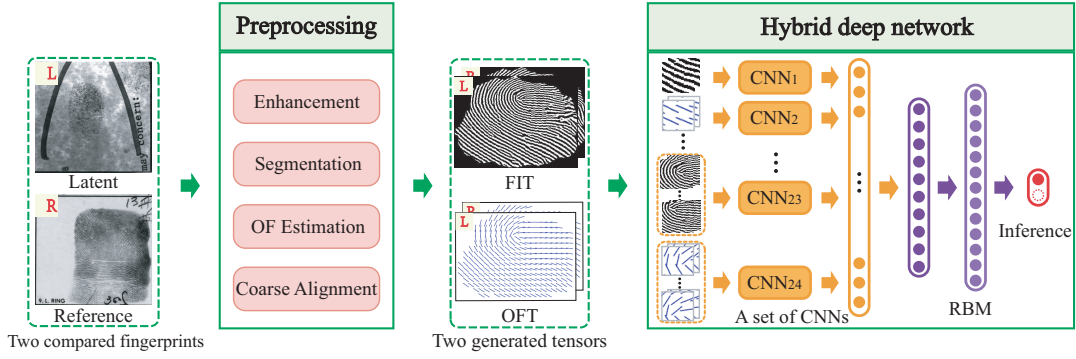}
	\vspace{-0.17in}
	\caption{Framework of the proposed latent fingerprint recognition method. L and R represent the latent and reference fingerprints, respectively. FIT and OFT are fingerprint image tensor and orientation field tensor, respectively. The set of CNNs is different in architectures and sizes (Section \ref{subsec:neuralnetworks}).}
	\label{fig:framework}
	\vspace{-0.1in}
\end{figure*}
Fig. \ref{fig:framework} illustrates the framework of the proposed method which consists of two main steps: 1) preprocessing two compared fingerprints to generate the required tensors and 2) pair-relationship modeling by a hybrid deep network. Specifically, for two compared fingerprints (a latent one and a reference one), their preprocessing includes enhancement, segmentation, orientation field (OF) estimation, and coarse alignment, which follows the usual preprocessing procedures in current common practice. The preprocessing generates two tensors: fingerprint image tensor (FIT) and orientation field tensor (OFT) which are concatenations of two preprocessed fingerprint images and their corresponding orientation fields, respectively (Section \ref{subsec:preprocessing}). The hybrid deep network, which is composed of a set of CNNs and an RBM, models the pair-relationship from multi-scale and multi-type patches of the FIT and OFT, and makes a prediction whether the input two fingerprints is a match (Section \ref{subsec:neuralnetworks}). The entire network is jointly trained and optimized (Section \ref{subsec:rbm}). 

\vspace{-0.1in}
\subsection{Preprocessing}
\label{subsec:preprocessing}
\begin{figure*}[b]
	\centering
	\includegraphics[width=0.78\linewidth]{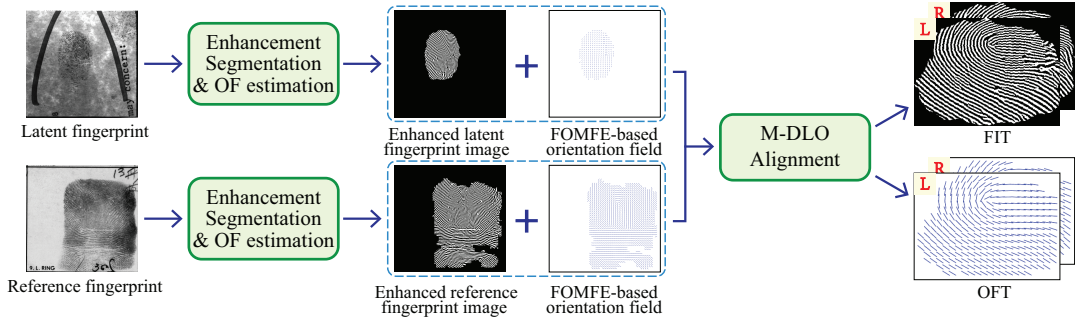}
	\vspace{-0.12in}
	\caption{Schematic diagram of the preprocessing. Note that the FIT and OFT can be concatenated in two ways, as detailed in Section \ref{subsub:FITgeneration}. Here, we only show one type of FITs and OFTs as an illustration.}
	\label{fig:preprocessing}
	%\vspace{-0.1in}
\end{figure*}
Fig. \ref{fig:preprocessing} illustrates the workflow of the preprocessing. For both a latent and a reference fingerprint, their enhanced fingerprint images and FOMFE-based orientation fields are firstly produced. Then, the M-DLO alignment is performed on the enhanced fingerprint images and the FOMFE-based orientation fields to generate the tensors FIT and OFT. Details are presented as follows.

\subsubsection{Enhanced Fingerprint Image and Corresponding Orientation Field Generation}
\label{subsub:enhancendFI}
Due to the different qualities of the latent and the reference fingerprints, we use different techniques to generate their enhanced fingerprint images and corresponding orientation fields. 
For the latent fingerprint, the method in Ref. \cite{21004dabouei2018id} is used to firstly produce an enhanced fingerprint image, an estimated orientation field, and a quality map. Then, the produced quality map is used to segment the region of interest (ROI) of the produced fingerprint image and orientation field. Finally, the segmented orientation field is further regularized by a fingerprint orientation model FOMFE \cite{wang2007fingerprint,RN759} for denoising and estimating orientations for moderately corrupted areas. 
For the reference fingerprint, because it is usually of good quality, the STFT method \cite{25chikkerur2007fingerprint} is used to produce its enhanced fingerprint image, estimated orientation field, and ROI segmentation. By the same token, its segmented orientation field is further regularized by the FOMFE model. Therefore, for both the latent and the reference fingerprints, their enhanced fingerprint images and FOMFE-based orientation fields are produced. 

\subsubsection{FIT and OFT Generation}
\label{subsub:FITgeneration}
To generate the tensors FIT and OFT, the two enhanced fingerprint images and the two FOMFE-based orientation fields need to be aligned first. Due to the poor quality of latent fingerprints, we propose to use orientation field for the alignment because it can provide a robust coarse alignment. We have observed that the random hexagon partition in the DLO alignment \cite{yager2005coarse} could generate significant discrepant hexagons over the ROI, leading to poor alignment results. To address this issue, we propose an improved minutia-based DLO alignment named as M-DLO alignment where the hexagons are formed around the most reliable minutia. One or two reliable minutiae will be sufficient to help produce a sound alignment with our proposed M-DLO alignment. Algorithm \ref{alg_RVCE} summarizes the M-DLO alignment. 
\begin{savenotes}
	\begin{algorithm}[ht]
		\caption{M-DLO alignment algorithm}
		\label{alg_RVCE}
		\begin{algorithmic}[1]
			\renewcommand{\algorithmicrequire}{\textbf{Input:}}
			\REQUIRE ~~\\
			Two fingerprints' FOMFE-based orientation fields ($P, Q$) and the corresponding two sets of minutiae\footnote{They are extracted on the enhanced fingerprint images using the method in \cite{RN252}.}.
			\renewcommand{\algorithmicensure}{\textbf{Output:}}
			\ENSURE ~~\\
			Geometrical transformation parameters ($\Delta{x}, \Delta{y},$ $\Delta{\theta}$)\footnote{($\Delta{x}, \Delta{y})$ are the location transformations and $\Delta{\theta}$ is the rotation transformation of the alignment.}.
			\renewcommand{\algorithmicensure}{\textbf{Operations:}}
			\ENSURE ~~\\
			\FOR{each fingerprint}
			\FOR{each of its minutiae}
			\STATE Take it as the center of a hexagon\footnote{The side length of the hexagon is $12$ pixels.} to partition this fingerprint's FOMFE-based orientation field into hexagonal elements, and calculate an average orientation\footnote{The average orientation is defined as the average of the orientations of each pixel within the element.} for each element. 
			\STATE Calculate a feature vector\footnote{The feature vector is defined as the relative orientations of this element and each of its six neighboring elements.} for the element centered at this minutia. 
			\ENDFOR
			\ENDFOR
			\STATE Find similar feature vectors by comparing each pair of feature vectors from the two orientation fields\footnote{The similarity of two feature vectors is calculated based on Euclidean distance. Top $20\%$ pairs of feature vectors are selected as similar ones.}.
			\STATE Get the corresponding pairs of minutiae of these similar feature vectors.
			\FOR {each obtained pair of minutiae}
			\STATE Align the two orientation fields using them and calculate the cost\footnote{The cost is calculated by $C(P,Q')=\sum_{p,q'}\lambda(p_\theta, q'_\theta)/N $ where $Q'$ is the obtained OF by applying transformation on $Q$, $p \in P$ and $q' \in Q'$ are the elements, $p_\theta$ and $q'_\theta$ are the average orientation of $p$ and $q'$, and $N$ is the number of overlapping element pairs.}.
			\ENDFOR
			\STATE Select the alignment leading to the smallest cost.
		\end{algorithmic}
	\end{algorithm}
\end{savenotes}

After aligning the two enhanced fingerprints and the two FOMFE-based orientation fields, they are respectively concatenated to generate the tensors FIT and OFT. There are two ways of the concatenation: 1) the fingerprint image (or orientation field) of the latent fingerprint at front and the one of reference fingerprint at back; and 2) in reverse. We name the tensors generated by these two ways as FIT$_1$ (or OFT$_1$) and FIT$_2$ (or OFT$_2$), respectively.

\subsection{Proposed Hybrid Deep Network}
\label{subsec:neuralnetworks}
In this section, an overview of the hybrid deep network is firstly presented in Section \ref{subsubsection:overview}. Then, the details of the set of CNNs are described in Section \ref{subsubsec:cCNN} and Section \ref{subsubsec:pCNN}, followed by the description of the RBM in Section \ref{subsubsec:RBM}. 

\subsubsection{Overview}
\label{subsubsection:overview}
For convenience, we define the architecture of a network only by the types of its layers and their connections, such that a network refers to an architecture with specific hyper-parameters, and a model refers to a network with specific inputs. 

\begin{figure*}[htb]
	\centering
	\includegraphics[width=\linewidth]{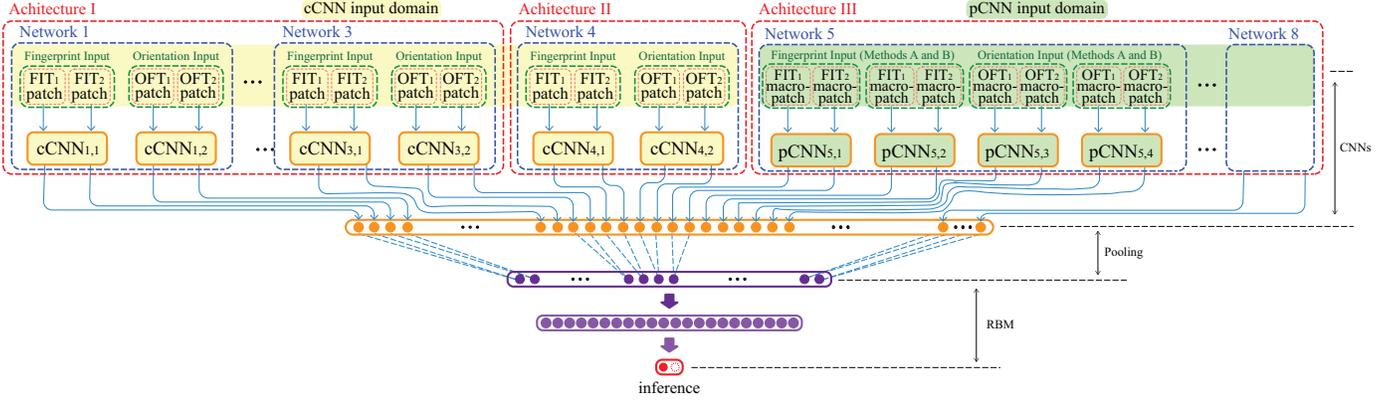}
	\vspace{-0.25in}
	\caption{Schematic diagram of the proposed hybrid deep network. It contains 24 CNN models which belong to eight networks and have only three architectures. The 24 CNN models are classified into two categories, namely cCNN and pCNN. The first subscript of the CNN models corresponds to the number of the network, and the second subscript corresponds to the index of the model under the same network. The inputs to different CNN models are different in sizes and types, as detailed in Sections \ref{subsubsec:cCNN} and \ref{subsubsec:pCNN}.}
	\label{fig:hybridnetwrok}
	\vspace{-0.05in}
\end{figure*}

\begin{figure}[t]
	\centering
	\includegraphics[width=\linewidth]{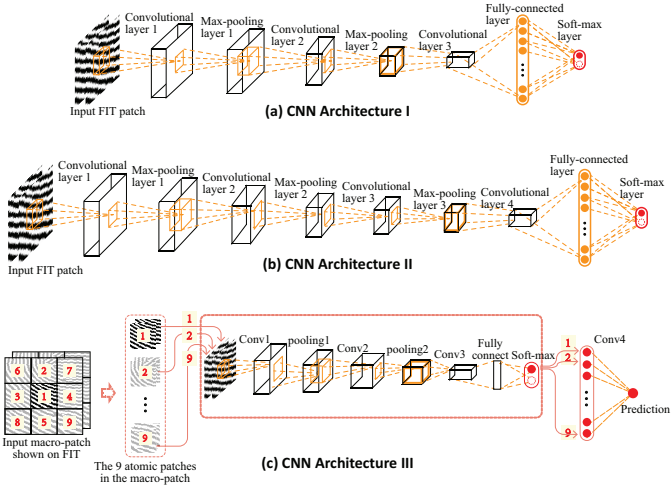}
	\vspace{-0.25in}
	\caption{Three architectures used in the proposed hybrid deep network. The black cuboids represent the feature maps and the orange cuboids represent the kernels of the convolutional layers and pooling layers. The inputs to architecture I and II are patches taken from tensors FIT and OFT. The inputs to architecture III are macro-patches taken from tensors FIT and OFT. Details are in Section \ref{subsubsec:cCNN} and \ref{subsubsec:pCNN}.}
	\vspace{-0.2in}
	\label{fig:cnn}
\end{figure}

Fig. \ref{fig:hybridnetwrok} illustrates the proposed hybrid deep network. Its inputs are patches (or macro-patches as detailed in Section \ref{subsubsec:pCNN}) taken from the FIT and OFT tensors, thus they are also tensors. In the remaining paper, we refer these patch-level tensors as patches (or macro-patches) unless stated otherwise. It have eight different CNN networks, which are constructed from three architectures I, II, and III, as illustrated in Fig. \ref{fig:cnn}. Architectures I and II are designed to extract features from multi-scale patches and architecture III is designed to learn the neighborhood relationship of a set of atomic patches in a macro-patch. They provide complementary information. Compared to architecture I, architecture II has one more convolution layer and one more pooling layer due to its larger input size. Compared with architectures I and II, architecture III has an additional convolution layer after the softmax for combining the softmax outputs of multiple input atomic patches. We name the CNN networks based on the architectures I and II as cCNNs, and the CNN networks based on the architecture III as pCNNs. 

\textit{cCNNs:} There are four cCNN networks corresponding to four input sizes 64$\times$64, 80$\times$80, 96$\times$96, and 192$\times$192 (detailed in Appendix A). Under each network, two models are trained to extract complementary information from two types of inputs: FIT type input and OFT type input. Hence, there are a total of eight cCNN models. Each model will generate two outputs corresponding to two sets of input patches cropped from FIT$_1$ and FIT$_2$ (or OFT$_1$ and OFT$_2$), respectively\footnote{The input patches to the cCNN models are obtained using a sliding window and thus there are many cropped patches corresponding to a cCNN model. They are all input to the corresponding cCNN model and their outputs are averaged for one output. Thus, for each model, there are two outputs corresponding to two sets of input patches cropped from FIT$_1$ and FIT$_2$ (or OFT$_1$ and OFT$_2$). \label{foot1}}. Therefore, there are 16 outputs from cCNNs.

\textit{pCNNs:} There are also four pCNN networks corresponding to four input sizes 32$\times$32, 48$\times$48, 64$\times$64, and 96$\times$96 (detailed in Appendix B). However, four models are trained under each network to extract complementary information from four types of inputs: FIT type input and OFT type input each generated through two methods $A$ and $B$ (detailed in Section \ref{subsubsec:pCNN}). Hence, there are a total of 16 pCNN models. Similar to cCNNs' output generation, there are 32 outputs from pCNNs.

A total of 48 outputs are generated by these 24 CNN models, and they pass through a pooling layer which can reduce the prediction variance by averaging the two outputs of each CNN model. Finally, the RBM makes a decision based on the outputs of this pooling layer. 

\textbf{Motivation:} Latent fingerprints have major challenges of distorted/corrupted areas and small ROIs. Our hybrid deep network is delicately designed to handle these challenges through the utilization of noise resistant orientation field, multi-scale patches, and neighborhood relationship of macro-patches. The patch partition can help generate some quality inputs to the network and reduce the impact of poor-quality parts due to the distorted/corrupted areas. As the average fingerprint ridge frequency (distance between ridges) is about eight pixels for images with 500dpi resolution \cite{27orczyk2011fingerprint}, the smallest atomic patch is set as 32$\times$32. This would make sure the atomic patch contains at least three ridges/valleys. A smaller size patch will likely contain one or two ridges/valleys, which is less reliable or discriminative. By the same token, the stride of different patch sizes is set to be 16 pixels, which can normally add two new ridges/valleys incrementally. Also, we have surveyed the sizes of the ROI area of all latent fingerprints in SD27 database which is a practical database provided by the FBI and found that the smallest ROI area is larger than the size of 32$\times$32 which fits well with the smallest atomic patch size we set. Accordingly, we have the patch size list 32$\times$32, 48$\times$48, 64$\times$64, 80$\times$80, 96$\times$96, $\dots$, 192$\times$192. The largest patch size is set to be 192$\times$192 because 95\% of the latent fingerprints have a ROI area larger than the size of 192$\times$192. A size larger than 192$\times$192 would increase the number of latent fingerprints whose ROI sizes are smaller than 192$\times$192, resulting patches with many void areas. The setting of the numbers and sizes of cCNNs and pCNNs in our deep network can systematically cover almost all the sizes in the above patch list. 

\subsubsection{Details of the cCNNs} 
\label{subsubsec:cCNN}
The cCNNs are designed to model the pair-relationship from patches. They have different input sizes and thus can encode multi-level local relationships into the pair-relationship. To enhance the network's discriminative power and reduce the number of parameters, we set the kernel size to be 3$\times$3 for all convolution layers except the last one. The kernel size of the last convolution layer is adaptively decided to ensure that the final output is always of size 1$\times$1. Therefore, the cCNN network corresponding to the input size of 192$\times$192 uses architecture II instead of architecture I (Details of the parameters of the four cCNN networks can be found in Appendix A). A cCNN network takes its corresponding sized patch as input and outputs a two-way softmax which represents the probability distribution over two classes. For all convolution layers except the last one, ReLU active function is adopted. Also, the dropout is employed as a regularizer on the fully-connected layer to prevent over-fitting. 

\textit{Training Input Generation:}
The input patches are cropped from the FIT and OFT tensors using a sliding window. In addition, the entire FIT and OFT tensors are resized to 192$\times$192 for use as patches, which ensures that the entire overlap of the two fingerprints is considered. To exclude unreliable patches from these patches, a quality evaluation needs to be conducted. As the FOMFE-based orientation is comparatively reliable, the quality evaluation is not conducted on the patches cropped from the OFT tensor. This also enables the OFT patches to provide complementary information and guidance in the case of missing quality FIT patches. The quality evaluation on the FIT patches is conducted as follows. Firstly, good-quality masks are respectively obtained for the latent and rolled fingerprints in the FIT. Then, only the patch with more than 75\% of its area being the overlap of two good-quality masks is treated as quality patch and kept. The good-quality mask for the latent fingerprint in the FIT is its quality map generated in Section \ref{subsub:enhancendFI}. The good-quality mask for the reference fingerprint in the FIT is generated as follows. Firstly, its quality map $\mathcal{Q}$ is obtained by evaluating its orientation coherence \cite{2MaltoniHandbook} which is formulated as:
\begin{equation}
\mathcal{Q} = \frac{\sqrt{(G_{x_0x_0} - G_{y_0y_0})^2+4G^2_{x_0y_0}}}{G_{x_0x_0} + G_{y_0y_0}},
\end{equation} 
with
\begin{equation}
\left\{ 
\begin{aligned}
G_{x_0x_0} & = \sum_{x\in \mathcal{N}} g_{x}^2 \\
G_{y_0y_0} &= \sum_{{y\in \mathcal{N}}} g_y^2  ~~~~~~~~~,\\
G_{x_0y_0} &= \sum_{{x,y\in \mathcal{N}}} g_xg_y \\
\end{aligned}
\right.
\end{equation}
where $(x_0, y_0)$ represents a 17$\times$17 window center. The local gradient is calculated as $(g_{x}, g_{y})$ in a 3$\times$3 window with $(x,y)$ being its center. $\mathcal{N}$ is a 17$\times$17 window centered at the pixel $(x_0, y_0)$. Then, the good-quality mask is generated by applying a threshold 0.9 on $\mathcal{Q}$.

\subsubsection{Details of the pCNNs}
\label{subsubsec:pCNN} 
The pCNNs are designed to model the pair-relationship from macro-patches which can be generated by two methods $A$ and $B$, and formulated as:
\begin{equation}
\begin{cases}
\Phi(A,j) = \{\phi_{c}(j)|c=(x+d_{x},y+d_{y}), d_{x},d_{y} \in \{\pm16,0\}\},
\\
\Phi(B,j) = \{\phi_{c}(j)|c=(x+d_{x},y+d_{y}), d_{x},d_{y} \in \{\pm32,0\}\},
\end{cases}
\label{equ:macro-patch}
\end{equation}
where $\Phi(A,j)$ and $\Phi(B,j)$ are macro-patches generated by the methods $A$ and $B$, respectively. $\phi_{c}(j)$ is one of the nine atomic patches, $j \in \{32,48,64,96\}$ represents the atomic patch size, and $c$ is the center coordinate of $\phi_{c}(j)$. $d_{x}$ and $d_{y}$ are the Euclidean distances along the $x-$ and $y-$ directions. This way, the macro-patch pair-relationship can be modeled by summarizing its nine atomic patches' pair-relationships through convolution. Such summarization can increase the robustness of the pair-relationship modeling against the unreliable pair-relationship occurred at some atomic patches.

All four pCNN networks use the same architecture III, which consists of a sub-architecture (marked by the pink dotted rectangle in the middle) identical to the architecture I and a following convolutional layer used to summarize the softmax outputs of nine input atomic patches, as shown in Fig. \ref{fig:cnn}(c). Detailed parameters of these four pCNN networks can be found in Appendix B. The inputs of the pCNNs are nine atomic patches in a macro-patch and the output is a prediction representing the pair-relationship of the macro-patch area. Hence, a pCNN can capture the neighborhood relationship of the central atomic patch and provide complementary information to the cCNNs. 

\textit{Training Input Generation:} The input macro-patches are cropped from the FIT and OFT tensors using a sliding window. Similar to the cCNN training input generation, the quality evaluation is also conducted only on the macro-patches cropped from the FIT to exclude unreliable ones. Also, the same good-quality masks used in the cCNN training input generation are used here. A FIT macro-patch will be kept if it has more than four atomic patches where each one has more than 75\% of its area inside the overlap of the two good-quality masks.

\subsubsection{RBM}
\label{subsubsec:RBM}
RBM is a generative stochastic artificial neural network which learns a probability distribution over its set of inputs. It has been successfully applied to a classification task by Hinton et al. \cite{34hinton2006fast}, where the joint distribution of an input $\textbf{x}=({x_1,\ldots,x_m})$ and output $y\in({1,\ldots,C})$ is modeled using a hidden layer of binary stochastic units $\textbf{h}=({h_1,\ldots,h_H})$. This is achieved by firstly defining an energy function as:
\begin{equation}
E(y,\textbf{x},\textbf{h})=-\textbf{h}^T\textbf{W}\textbf{x}-\textbf{r}^T\textbf{x}-\textbf{s}^T\textbf{h}-\textbf{t}^T\textbf{e}_y-\textbf{h}^T\textbf{U}\textbf{e}_y,
\end{equation}
where $\textbf{W}$ and $\textbf{U}$ are weight matrices. $\textbf{r}$, $\textbf{s}$, and $\textbf{t}$ represent the bias weights (offsets) for input $\textbf{x}$, hidden units $\textbf{h}$ and the representation $\textbf{e}_y$, respectively. $\textbf{e}_y=(1_{i=y})^C_{i=1}$ is the `one out of $C$' representation of $y$. Then, the probabilities of the values of $y$ and $\textbf{x}$ can be assigned as \cite{34hinton2006fast}:
\begin{equation}
\label{equ:8}
\begin{aligned}
&p(y|\textbf{x}) = \\ &\frac{\exp(t_y+\sum_jg(s_j+U_{jy}+\sum_iW_{ji} x_i))}{\sum_{y*\in\{1,\ldots,C\}}\exp(t_{y*}+\sum_jg(s_j+U_{jy*}+\sum_i W_{ji}x_i))},
\end{aligned}
\end{equation}
where $g(\cdot)=\log(1+\exp(\cdot))$. 

The RBM is adopted in the proposed hybrid deep network to combine the outputs of the set of CNNs and make a decision on whether an input fingerprint pair is a genuine pair or not. Since it is conducted in a supervised learning setting and only needs to obtain a good prediction of the target according to the given input, discriminative training is adopted to minimize the objective function $\mathcal{L}$ formulated as:
\begin{equation}
\mathcal{L}(\mathcal{T})=
\sum_{i=1}^{|\mathcal{T}|}L(\textbf{x}_i),
\end{equation}
where $\mathcal{T}={(\textbf{x}_i, y_i)}$ represents the training set, and $L(\textbf{x}_i)=-\log p(y_i|\textbf{x}_i)$ is the RBM loss function. 

\subsection{Joint Optimization}
\label{subsec:rbm}
The hybrid deep network is jointly optimized as follows. Firstly, the set of CNNs are trained separately using the binary cross entropy loss, where the binary classes are the binary fingerprint verification targets. The loss is minimized by stochastic gradient descent, where the gradient is calculated by back-propagation. Then, the RBM is trained by fixing the CNNs and its loss is also optimized by stochastic gradient descent as the CNNs. Its gradient $-\frac{\partial logp(y|\textbf{x})}{\partial \theta } $ can be computed explicitly due to the closed form expression of the likelihood, where $\theta \in \{W, U, s, t\}$ are parameters
to be learned. Finally, the entire network is tuned by the back-propagating errors from the RBM layer to all the CNNs' layers with the gradient of the loss with regard to $w_q$, which is calculated by:
\begin{equation}
\frac{\partial L}{\partial w_q}=\frac{\partial L}{\partial x_q}\frac{\partial x_q}{\partial w_q},
\end{equation} 
where $w_q$ and $x_q$ are the weight value and prediction of the $q$-th CNN, respectively. $\frac{\partial L}{\partial x_q}$ is calculated by the closed form expression of Eq. (\ref{equ:8}) and $\frac{\partial x_q}{\partial w_q}$ can be calculated using
the back-propagation algorithm in the CNNs.

\section{Experimental Results and Evaluation}
\label{sec:experimentalresults}
In this section, experiment settings are firstly detailed in Section \ref{subsec:setting}. Then, the performance evaluation and comparison are presented in Section \ref{subsec:PEC}, followed by the analysis and robustness validation of the proposed method in Section \ref{subsec:discussion}.

\subsection{Experiment Settings}
\label{subsec:setting}
\subsubsection{Training Database}
A challenging problem of applying deep learning to latent fingerprint application is the training database preparation. As the current public databases are either short of the correspondence between latent fingerprints and their true mates or lack of quantity, they are more suitable to be used as the evaluation databases to evaluate the performance of the proposed method. Therefore, for training the hybrid deep network, we propose to generate the training database using the NIST SD14 database \cite{37watson2001nist} as follows. 

The NIST SD14 database consists of 54,000 rolled fingerprints, and thus we synthesize 54,000 corresponding latent fingerprints for them. To generate latent fingerprints that better mimic real cases, we propose to add complex and realistic noise to the rolled fingerprints to synthesize their latent ones. This also enables the proposed deep network to learn more effective pair-relationship modeling from tough latent fingerprint situations.

\begin{figure}[h]
	\centering
	\includegraphics[width=0.8\linewidth]{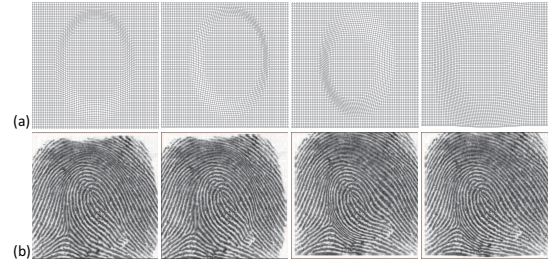}
	\vspace{-0.1in}
	\caption{Samples of (a) four generated plastic distortions and (b) their related distorted fingerprints for one rolled fingerprint.}
	\label{fig:distortion}
\end{figure}

\begin{figure}[h]
	\centering
	\includegraphics[width=0.8\linewidth]{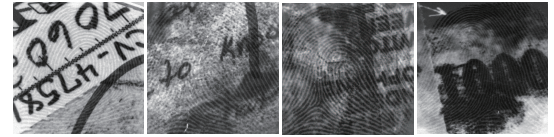}
	\vspace{-0.1in}
	\caption{Four examples of the synthesized latent fingerprints.}
	\label{fig:latent}
\end{figure}

To this end, firstly, a realistic noise image database is created, which consists of noise images with diverse types of noises. Those noise images are cropped from natural images and resized to the same size as the rolled fingerprint images in NIST SD14 database. Then, given a rolled fingerprint $I$, a plastic distortion \cite{41cappelli2001modelling} is added by
\begin{equation}
\textbf{v}' = \textbf{v} +\Delta (\textbf{v})\cdot g(f(\textbf{v}),l),
\end{equation}
where $\textbf{v} = [x, y]^T$ is a point in $I$ and $\textbf{v}' $ is its distorted point. $l$ is the skin plasticity coefficient. $\Delta (\textbf{v})$ is the torsion and traction amount computed on the basis of a rotation angle $\theta$ and a displacement vector $\textbf{a} = [a_x, a_y]^T$, which is given by
\begin{equation}
\Delta (\textbf{v}) = (\textbf{R}_{\theta}(\textbf{v}-\textbf{o}_r)+\textbf{o}_r+\textbf{a})-\textbf{v},
\end{equation}
\begin{equation}
\textbf{R}_{\theta} = \begin{bmatrix}
cos \theta & sin\theta \\ 
-sin\theta & cos\theta 
\end{bmatrix},
\end{equation}
where $\textbf{o}_r$ is the center of rotation. $g(f(\textbf{v}),l)$ is the gradual transition defined as:
\begin{equation}
g(f(\textbf{v}),l) = \begin{cases}
0, ~~~~~~~~~~~~~~~~~~~~~~~~~~~~~ f(\textbf{v}) < 0\\ 
\frac{1}{2}(1 - cos(\frac{\pi \cdot f(\textbf{v})}{l})),  ~~~~~~0< f(\textbf{v}) < l\\ 
1,  ~~~~~~~~~~~~~~~~~~~~~~~~~~~~~otherwise.
\end{cases}
\end{equation}
Function $f(\textbf{v})$ returns a measure proportional to the distance between the point and the border of an ellipse centered at $\textbf{o}_e$ with semi-axes $s_x$ and $s_y$, and is formulated as:
\begin{equation}
f(\textbf{v}) = \sqrt{(\textbf{v}-\textbf{o}_e)^T\textbf{A}^{-1}(\textbf{v}-\textbf{o}_e)}-1,
\end{equation}
\begin{equation}
\textbf{A} = \begin{bmatrix}
s^2_x & 0 \\ 
0 & s^2_y
\end{bmatrix}.
\end{equation} 
These plastic distortion parameters $l$, $\theta$, $(a_x, a_y)$, $s_x$, and $s_y$ are set in the ranges of $[0.5, 2]$, $[0, 5]$, $[-15, 15]$, $[0.2b, 0.6b]$, and $[0.2b, 1.2b]$, respectively, where $b$ is half the width of the image. Fig. \ref{fig:distortion} illustrates samples of multiple generated distortions and their related distorted fingerprints for one rolled fingerprint. 
Finally, a synthesized latent fingerprint is generated by integrating the distorted fingerprint and a noise image randomly picked up from the created noise image database, with the intensity degree of the noise image ranging from $0.2$ to $0.8$. Fig. \ref{fig:latent} shows some examples of the synthesized latent fingerprints. Overall, the training database consists of the rolled fingerprints in the NIST SD14 database and their generated corresponding latent fingerprints.

\subsubsection{Training Settings}
\label{subsub:TS}
In the NIST SD14 database, the two fingerprint impressions of each finger are stored with labels `f' and `s'. A positive pair of fingerprints is formed by the latent fingerprint derived from `f' and the rolled fingerprint `s', or the latent fingerprint derived from `s' and the rolled fingerprint `f'. Thus, there is a total of 54,000 positive pairs of fingerprints for the training. The negative pairs of fingerprints are formed by randomly selecting a non-mated rolled fingerprint for each of the synthesized latent ones, and thus the total number of the negative pairs is also 54,000. We randomly choose 80\% of these pairs of fingerprints to train the set of CNNs and use the remaining 20\% pairs to train the RBM and tune the entire network.

\subsubsection{Evaluation Databases}
\textit{Latent fingerprint databases:} In line with current methods, the latent fingerprint databases NIST SD27 \cite{22NIST} and IIIT-Delhi MOLF \cite{sankaran2015multisensor} are used to evaluate the proposed method.

\textit{Reference fingerprint databases:} In the literature, many methods use different reference databases. For a comprehensive evaluation, four reference databases are used in this paper and summarized as follows.

\textit{Reference database DB-A}: It consists of the rolled fingerprints provided in NIST SD27 database.

\textit{Reference database DB-B}: It consists of the rolled fingerprints provided in NIST SD27 database and the 2,000 fingerprints in NIST SD04 database.

\textit{Reference database DB-C}: It consists of the rolled fingerprints provided in NIST SD27 database and the 27,000 fingerprints in NIST SD14 database.

\textit{Reference database DB-L}: In general, the larger the reference database, the closer it will be to the realistic application of latent fingerprint recognition. The latest methods \cite{21cao2018automated} \cite{cao2018end} use a large-scale reference database consisting of 100,000 fingerprints. However, it is not publicly available. Therefore, the large-scale reference database DB-L is created using public rolled fingerprint databases: NIST SD04 \cite{36watson1992nist}, NIST SD14 \cite{37watson2001nist}, FVC2002 DB1A-DB4A \cite{RN753}, FVC2004 DB1A-DB4A \cite{RN2022}, FVC2006 DB1A and DB3A \cite{RN310}, Sokoto \cite{shehu2018detection}, the database in \cite{RN440}, the database in \cite{puri2010analysis}, and the database in \cite{lin2018matching}, and the associated full fingerprints in databases: NIST SD27 \cite{22NIST}, IIIT-Delhi MOLF \cite{sankaran2015multisensor}, and IIIT-D Latent \cite{sankaran2011matching}, which has a total of 97,998 fingerprints. All the fingerprint images are resized to 500ppi for the recognition. Note that the NIST SD14 database has been only used for synthesizing latent training data and has not been used in any way for the recognition task, and thus it can be used here to enlarge the reference database. 

\subsubsection{Evaluation Settings}
\label{subsub:ES}
For the evaluation, all patches and macro-patches cropped from the FIT and OFT are correspondingly input to the trained hybrid deep network. The quality evaluation used in the training input generation is not used here. For the case of the size of the FIT and OFT being smaller than the size of the required patch (or macro-patch), the required patch (or macro-patch) is generated by filling the remaining area with $255$. The performance of the latest method \cite{cao2018end} reported in the following is obtained by running its public code.

\subsection{Performance Evaluation and Comparison}
\label{subsec:PEC}
\subsubsection{Evaluation on NIST SD27 Database }
\label{subsec:comparison}
The NIST SD27 database is provided by the NIST in collaboration with the FBI. It consists of 258 crime-scene latent fingerprints, which are classified based on three different image qualities, namely `good', `bad', and `ugly', with numbers of images 88, 85, and 85, respectively. The resolutions of these fingerprint images are 500ppi. Many methods \cite{8jain2011latent,10jain2008matching,11001arora2014latent,17yoon2011latent,18paulino2013latent,19001medina2016latent,21cao2018automated,cao2018latent,cao2018end} have been evaluated on this database but their reference databases vary. We evaluated the proposed method (using DLO and M-DLO alignments respectively) over the four reference databases (DB-A, DB-B, DB-C, DB-L). Table \ref{table:0000} compares the rank-1 accuracies of the proposed method, the recent method \cite{19001medina2016latent} which uses manually marked minutiae, and the latest method \cite{cao2018end}. It has been reported that using manually marked minutiae could improve the accuracy performance by up to 38\% \cite{16yoon2010latent}. This is not surprising because it gets the benefit of the assistance from latent fingerprint experts.
Compared with these results achieved using manually marked minutiae, our results without using any manual inputs are competitive, with two results being very close and one result being better. Also, our results evaluated using four reference databases are all superior to those of the latest method \cite{cao2018end}. 

\begin{table}[htbp]
	\renewcommand{\arraystretch}{1.7}
	\caption{Comparison of rank-1 accuracies (\%) of different methods evaluated on NIST SD27 database using the reference databases DB-A, DB-B, DB-C, and DB-L} 
	\vspace{-0.25in}
	\label{table:0000}
	\begin{center}
		\begin{tabular}{c|c|c|c|c}		
			\Xhline{0.9pt}	
		    & Medina\cite{19001medina2016latent}* & Cao\cite{cao2018end} & \tabincell{c}{Our Method\\ (with DLO)} & \tabincell{c}{Our Method\\ (with M-DLO)}\\
			\Xhline{0.9pt}	
			\textbf{DB-A} & 83.9 & 79.1 & 77.9& 81.4\\ 
			\hline
			\textbf{DB-B} & 76.9 & 74.8 & 76.0 & 76.7\\
			\hline
			\textbf{DB-C} & 67.3 & 70.2 & 71.3 &71.7\\
			\hline	
			\textbf{DB-L} & N/A & 66.3 & 65.9 &67.8\\
			\Xhline{0.9pt}
		\end{tabular}
		\begin{tablenotes}
			\footnotesize
			*: This method uses manually marked minutiae and is used as a baseline. 
		\end{tablenotes}
	\end{center}
	\vspace{-0.1in}
\end{table}

\begin{figure}[h]
	\centering
	\includegraphics[width=0.8\linewidth]{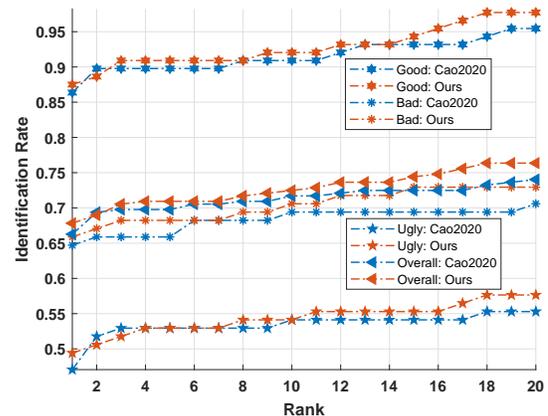}
	\vspace{-0.12in}
	\caption{CMC curves of different methods evaluated on the NIST SD27 database using the large-scale reference dataset DB-L with respect to both the overall and the three different quality types (`good', `bad', and `ugly') of latent fingerprints.}
	\label{fig:roc1}
\end{figure}

Fig. \ref{fig:roc1} compares the CMC curves of the proposed method and the latest method \cite{cao2018end} (labeled as Cao2020 in the figure) evaluated using the reference database DB-L with respect to both the overall and the three quality types of latent fingerprints in the NIST SD27 database. As can be seen, the performance of all algorithms for good latent fingerprints are significantly better than those for bad and ugly latent fingerprints. This is expected because the quality of fingerprints has an important impact on the recognition performance. The low-quality latent fingerprints usually mean seriously corrupted ridges and small regions of fingers, which increase the difficulty of recognition. The proposed method achieves better results than the latest method on all these three quality types of fingerprints. 

\subsubsection{Evaluation on IIIT-Delhi MOLF Database}
\label{subsec:more}
\begin{figure}[ht]
	\centering
	\includegraphics[width=0.8\linewidth]{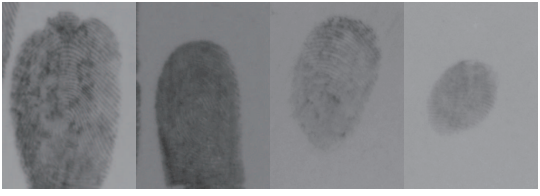}
	\vspace{-0.1in}
	\caption{Samples of latent fingerprints from IIIT-Delhi MOLF database.}
	\label{fig:IIIT}
	\vspace{-0.05in}
\end{figure}

The IIIT-Delhi MOLF database is publicly provided by \cite{sankaran2015multisensor}. It contains 4,400 latent fingerprints sampled from 10 fingers of 100 individuals. The associated reference fingerprints are captured using three different sensors and partitioned into three sub-databases labeled as `L', `S', and `C', respectively. The resolution of these fingerprint images is 500ppi. These latent fingerprints are very challenging for recognition due to their bad quality, as shown in Fig. \ref{fig:IIIT}. The latent fingerprint recognition experiments are conducted according to the testing protocol established in Experiment III in \cite{sankaran2015multisensor}. The recognition performance reported in \cite{sankaran2015multisensor} is used as a baseline.

\begin{figure}[htbp]
	\centering
	\includegraphics[width=0.8\linewidth]{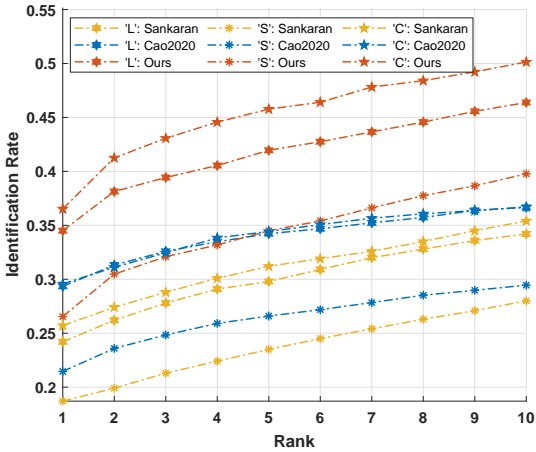}
	\vspace{-0.12in}
	\caption{CMC curves of different methods evaluated on the IIIT-Delhi MOLF database with respect to its three reference sub-datasets labeled as `L', `S', and `C'.} 
	\vspace{-0.1in}
	\label{fig:moreresults}
\end{figure}

Fig. \ref{fig:moreresults} compares the CMC curves of the proposed method, the latest method \cite{cao2018end} (labeled as Cao2020 in the figure), and the baseline with respect to the three reference sub-databases. As can be seen, the proposed method achieves much better rank-1 accuracy and a continuously rising identification rate with the increase of $N$ for all three reference sub-databases. This is attributed to: 1) the proposed pair-relationship modeling is robust for recognition because it does not rely on extracting reliable representation features from a latent fingerprint image; and 2) the preprocessing provides relatively reliable OFT as a supplement to FIT for the pair-relationship modeling.

\subsection{Ablation Experiments and Discussions}
\label{subsec:discussion}
\subsubsection{Architecture Variants Analysis}
\label{subsub:AVA}
To analyze the soundness of the proposed hybrid deep network, latent fingerprint recognition experiments using different architecture variants are conducted as follows. 

\textit{Experiment I}: half of the proposed 24 CNNs are used for the latent fingerprint recognition. This is achieved by removing a total of $12$ CNNs: the cCNN$_1$ (including cCNN$_{1,1}$ and cCNN$_{1,2}$ ), cCNN$_3$ (as cCNN$_1$), pCNN$_5$ (including pCNN$_{5,1}$, pCNN$_{5,2}$, pCNN$_{5,3}$ and pCNN$_{5,4}$), and pCNN$_7$ (as pCNN$_5$), and using the remaining 12 CNNs to form a new architecture which is named as M-Half.

\textit{Experiment II}: only cCNN-typed CNNs are used for the latent fingerprint recognition. Specifically, cCNN$_1$ (including cCNN$_{1,1}$ and cCNN$_{1,2}$), cCNN$_2$ (as cCNN$_1$), cCNN$_3$ (as cCNN$_1$), and cCNN$_4$ (as cCNN$_1$) are used to form a new architecture named as M-C.

\textit{Experiment III}: only CNNs taking FIT patches or marco-patches as input are used for the latent fingerprint recognition. Specifically, cCNN$_{1,1}$, cCNN$_{2,1}$, cCNN$_{3,1}$, cCNN$_{4,1}$, pCNN$_{5,1}$, pCNN$_{5,2}$, pCNN$_{6,1}$, pCNN$_{6,2}$, pCNN$_{7,1}$, pCNN$_{7,2}$, pCNN$_{8,1}$, and pCNN$_{8,2}$ are used to form a new architecture named as M-F.

\begin{figure}[ht]
	\centering
	\includegraphics[width=0.8\linewidth]{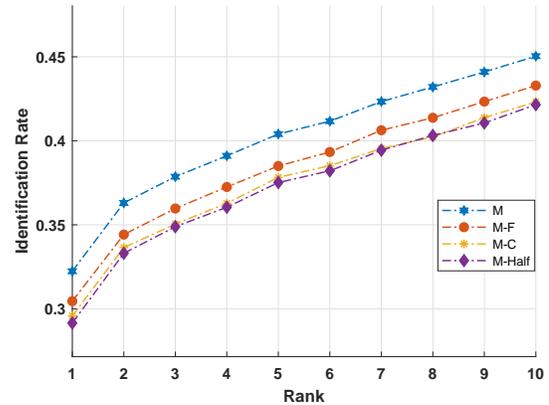}
	\vspace{-0.12in}
	\caption{CMC curves of the three models (M-half, M-C, M-F) and the proposed model (M) evaluated on the IIIT-Delhi MOLF database.}
	\label{fig:variants}
\end{figure}

\begin{figure}[ht]
\vspace{0.05in}
	\centering
	\includegraphics[width=0.8\linewidth]{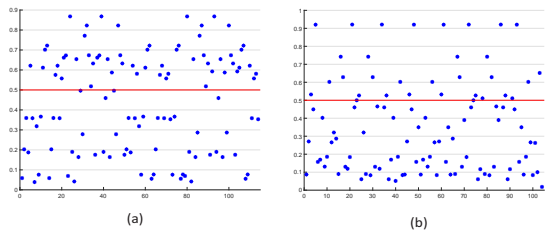}
	\vspace{-0.12in}
	\caption{The outputs of cCNN$_{1,1}$ from the model M for (a) a true pair of fingerprints and (b) a false pair of fingerprints.}
	\label{fig:points}
\end{figure}

These experiments are conducted using the experiment settings described in Section \ref{subsub:TS} and \ref{subsub:ES}. The performance of these three models is evaluated using IIIT-Delhi MOLF database with testing protocol described in Section \ref{subsec:more}. Fig. \ref{fig:variants} compares the CMC curves of these three models (M-Half, M-C, M-F) and the proposed full model (with $24$ CNNs and named as M). As can be seen, the performance of these three models is worse than that of the proposed full model. Specifically, first, the results of M-Half and M-C support our claim that different sizes and types of CNNs can model multi-level 
pair-relationship and provide complementary information to contribute to the final decision. This can be further explained by observing the outputs of the CNNs, which are removed in forming the models M-half, M-C, and M-F, in the model M. Fig. \ref{fig:points} shows the outputs (before average, see Footnote \ref{foot1}) of one of these CNNs (cCNN$_{1,1}$) for a true pair of fingerprints (a latent fingerprint and its true match) and a false pair of fingerprints. As can be seen, nearly half of the outputs (49\%) of the true pair are greater than 0.5, and most outputs (72\%) of false pair are less than 0.5. This proves the contribution of cCNN$_{1,1}$ (removed and not included in M-Half) to the final decision and explains why the result of M-half is worse than that of M. Second, the result of M-F demonstrates the necessity and effectiveness of incorporating the orientation field for the pair-relationship modeling.

\subsubsection{Analysis of Alignment Accuracy and Its Impacts on Recognition Performance}
\textit{Alignment Accuracy Analysis:} Two experiments IV and V are conducted to evaluate the alignment accuracy in terms of two metrics.

\begin{figure}[htbp]
	\centering
	\includegraphics[width=0.8\linewidth]{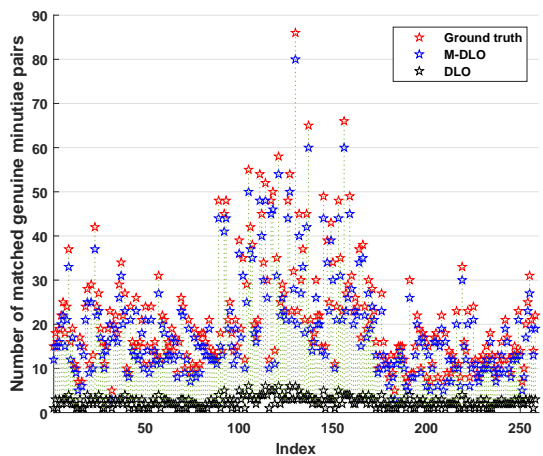}
	\vspace{-0.1in}
	\caption{Numbers of matched genuine minutiae pairs by the DLO alignment (black), the proposed M-DLO alignment (blue), and the manually marked ground-truth (red) for latent fingerprints in NIST SD27 database. The horizontal and vertical axes represent the latent fingerprint indexes and the number of matched genuine minutiae pairs, respectively. The mean-square error (MSE) of the number of matched genuine minutia pairs by the DLO and M-DLO compared with the ground-truth are 480.7519 and 7.3178, respectively. The average numbers of unmatched genuine minutia pairs by the DLO and M-DLO are 18.72 and 2.47, respectively.}
	\vspace{-0.1in}
	\label{fig:groundtruth}
\end{figure}

\textit{Experiment IV:}
\label{subsub:COAG}
The number of matched genuine minutia pairs after the proposed M-DLO alignment is compared with that of the DLO alignment and the manually marked minutiae. Fig.\ref{fig:groundtruth} shows the statistical results on the NIST SD27 database. As can be seen, the M-DLO alignment is significantly better than the DLO alignment with larger numbers of matched genuine minutia pairs and comparable with the ground-truth alignment with similar numbers of matched genuine minutia pairs. The MSE values 480.7519 and 7.3178 obtained by the D-LO and M-DLO alignments mean that the average numbers of unmatched genuine minutia pairs are 18.72 and 2.47, respectively. 

\begin{figure}[htbp]
	\centering
	\includegraphics[width=0.8\linewidth]{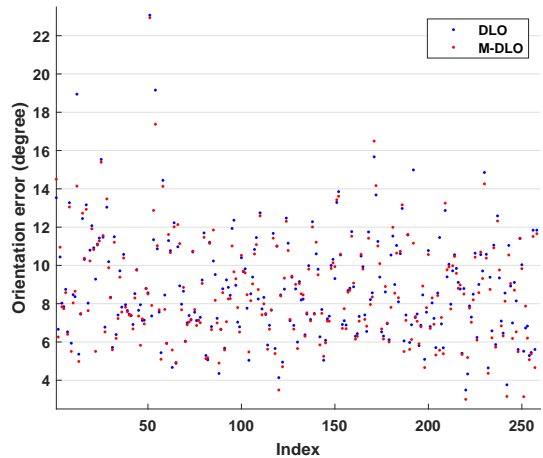}
	\vspace{-0.1in}
	\caption{Alignment accuracy, in terms of orientation error, of the DLO alignment (blue) and the proposed M-DLO alignment (red) for latent fingerprints in NIST SD27 database. The horizontal and vertical axes represent the latent fingerprint index and the difference of two orientations after alignment (latent against rolled), respectively. The mean-square error (MSE) of the orientation error of DLO and M-DLO are 84.38 and 80.99, respectively. The average differences between two aligned orientations by the DLO and M-DLO are 8.78 and 8.59, respectively.a}
	\label{fig:orientationalign}
\end{figure}

\textit{Experiment V:}
The orientation errors produced by the DLO and the proposed M-DLO alignments are compared. The orientation error measures the difference between two aligned orientation fields of the latent and its corresponding rolled fingerprints and is calculated using the Eq. (3) in method \cite{yager2005coarse}. Fig. \ref{fig:orientationalign} shows the statistical results on the NIST SD27 database. As can be seen, the M-DLO alignment achieves a moderately smaller alignment error than the DLO alignment. The MSE values 84.38 and 80.99 mean that the average differences between two aligned orientations by the DLO and M-DLO alignments are 8.78 and 8.59, respectively.

\textit{Alignment Impacts on Recognition Performance:}
Based on the experiments IV and V, we can conclude that: 1) the M-DLO alignment is more accurate than the DLO alignment. This alignment improvement also leads to the corresponding recognition performance improvement (see Table \ref{table:0000}). 2) The hexagon partition of the DLO algorithm is constructed around a random point which could produce inconsistent hexagon partitions. Our proposed M-DLO constructs the hexagon partitions based on the most reliable minutia, which leads to better consistent partitions evidenced by more matched genuine minutia pairs. 3) A larger number of unmatched pairs of minutiae will generally lead to a poor recognition performance in many minutia-based algorithms. However, our recognition network depends on the orientation field instead of direct minutia features. Both DLO and M-DLO alignments are orientation-based coarse alignments, but they also provide the advantage of noise resistance due to their averaging characteristics. Also, both the DLO and M-DLO alignments select the alignment point based on the minimum average orientation difference cost among all candidate points. Therefore, the minimum alignment error, in terms of orientation difference cost, is guaranteed even though there exists a certain number of unmatched minutiae. Therefore, our proposed recognition network is less sensitive to the number of matched genuine minutia pairs. This is evidenced by the graceful performance degradation (see Table \ref{table:0000}) even though there exists a significant variation of the number of matched genuine minutia pairs.   

\subsubsection{Preprocessing Impacts on Recognition Performance}
\begin{figure}[ht]
	\centering
	\includegraphics[width=0.8\linewidth]{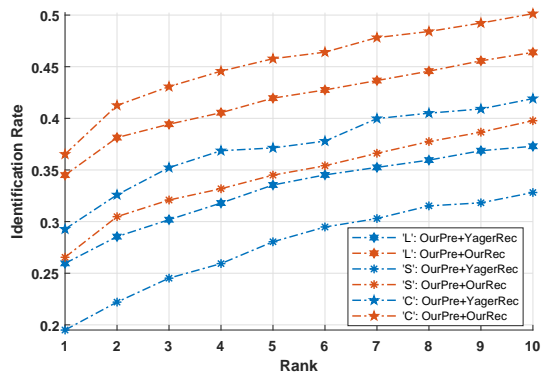}
	\vspace{-0.1in}
	\caption{CMC curves of the two different methods using the same preprocessing module (ours) and different recognition methods (Yager's \cite{yager2005coarse} and ours) evaluated on the IIIT-Delhi MOLF database with respect to its three reference sub-databases (`L',`S',`C').}
	\vspace{-0.05in}
	\label{fig:orientationmatch}
\end{figure}

Two ablation experiments VI and VII that using the same preprocessing module but different recognition methods are conducted to evaluate the preprocessing impact on the recognition performance. Results are summarized in Table \ref{table:0001}, and shown in Fig. \ref{fig:orientationmatch}.
\label{subsubsec:cross}

\begin{table}[htb]
	\renewcommand{\arraystretch}{1.4}
	\caption{Summary of the two ablation experiments VI and VII that using the same preprocessing module and different recognition methods} 
	\vspace{-0.25in}
	\label{table:0001}
	\begin{center}
		\begin{tabular}{c|c|c|c}		
			\Xhline{0.9pt}	
			& \tabincell{c}{Preprocessing \\Module} & \tabincell{c}{Recognition \\Method} & \tabincell{c}{Rank-1 \\Accuracy (\%)}\\
			\Xhline{0.9pt}	
			\multirow{2}{*}{\tabincell{c}{Experiment\\ VI}}& \tabincell{c}{Cao's (+ M-DLO)} & Ours & \tabincell{c}{68.22}\\ 
			\cline{2-4}
			& \tabincell{c}{Cao's} & Cao's \cite{cao2018end} & \tabincell{c}{66.27}\\
			\hline
			\hline
			\multirow{6}{*}{\tabincell{c}{Experiment \\VII}}& \multirow{3}{*}{\tabincell{c}{Ours}} & \multirow{3}{*}{Ours} & \tabincell{c}{34.50 (L)}\\ 
			\cline{4-4}
			&&&\tabincell{c}{26.55 (S)}\\ 
			\cline{4-4}
			&&&\tabincell{c}{36.52 (C)}\\ 
			\cline{2-4}
			& \multirow{3}{*}{\tabincell{c}{Ours}} & \multirow{3}{*}{Yager's \cite{yager2005coarse}} & \tabincell{c}{25.95 (L)}\\ 
			\cline{4-4}
			&&&\tabincell{c}{19.50 (S)}\\ 
			\cline{4-4}
			&&&\tabincell{c}{29.27 (C)}\\ 
			\Xhline{0.9pt}
		\end{tabular}
		\begin{tablenotes}
			\footnotesize
			The experiment VI is conducted on the NIST SD27 database. The experiment VII is conducted on the IIIT-Delhi MOLF database. The labels `L', `S', and `C' represent the three sub-datasets in the IIIT-Delhi MOLF database.
		\end{tablenotes}
	\end{center}
	\vspace{-0.1in}
\end{table}

\textit{Experiment VI:} The same preprocessing module (Cao's \cite{cao2018end}) with different recognition methods (Cao's and ours) are respectively used for the latent fingerprint recognition on the NIST SD27 database against the reference database DB-L.

\textit{Experiment VII:} The same preprocessing module (ours) with different recognition methods (Yager's \cite{yager2005coarse} and ours) are respectively used for the latent fingerprint recognition on the IIIT-Delhi MOLF database with respect to the three reference sub-databases `L', `S', and `C'. 

In experiment VI, Cao’s preprocessing module plus M-DLO alignment and Cao’s preprocessing module are regarded as the same preprocessing module because: 1) Cao’s system is independent of alignment. Hence, it implies that its system performance is independent of alignment, i.e., Cao’s preprocessing module plus M-DLO is virtually the same as using Cao’s preprocessing module in Cao's framework. 2) It is impossible to directly connect Cao’s preprocessing module to our recognition network if alignment is not added because our recognition network requires inputs of tensors of latent-rolled fingerprints where a coarse registration is required. Therefore, the best we can do is to use Cao’s preprocessing module plus the alignment component. Also, the enhanced fingerprint image generated by Cao's preprocessing method is a gray image, and we translate it into the binary image by a threshold $128$. The generated orientation field is also regularized by the FOMFE model for the alignment.

Table \ref{table:0001} summarizes the results of these two ablation experiments. As can be seen, in the case of using Cao's preprocessing method with different recognition methods, our proposed recognition method has enhanced the rank-1 accuracy by 2.94\%. In the case of using ours preprocessing method with different recognition methods, our recognition method has enhanced the rank-1 accuracy by 41.32\%, 40.64\%, and 40.86\% over sub-databases `L', `S', and `C', respectively. These experimental results strongly support the capability of our recognition network. 

\begin{figure}[htb]
	\centering
	\includegraphics[width=0.8\linewidth]{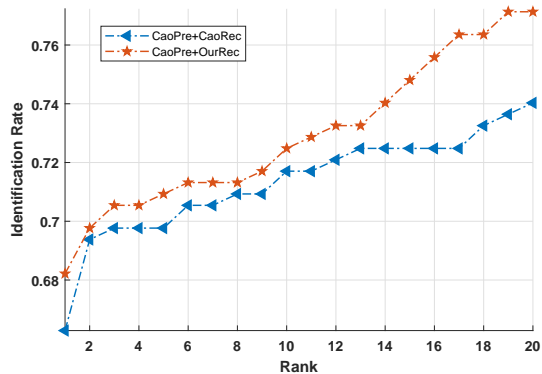}
	\vspace{-0.1in}
	\caption{CMC curves of the two different methods using the same preprocessing module (Cao's) and different recognition methods (Cao's \cite{cao2018end} and ours) evaluated on the NIST SD27 database.}
	\vspace{-0.05in}
	\label{fig:samePreanddiffRec}
\end{figure}
Fig. \ref{fig:samePreanddiffRec} and Fig. \ref{fig:orientationmatch} compares respectively the CMC curves of the two methods in experiment VI (labeled as `CaoPre+CaoRec' and `CaoPre+OurRec' in the figure) and the two method in experiment VII (labeled as `OurPre+YagerRec' and `OurPre+OurRec' in the figure). As can be seen, the performance achieved by using our recognition network is much better than using the other recognition methods, which demonstrates the soundness and robustness of the proposed hybrid deep network.

An illustration example (see Appendix D) is also provided as follows. For a pair of fingerprints recognized by the proposed recognition method at rank-1, an alignment noise $(\Delta{x} =10,\Delta{y}=-10,\Delta{\theta}=10^{\circ})$ is added to its M-DLO alignment to generate new FIT and OFT tensors input to the hybrid deep network for the recognition. Experimental result shows that it can still be recognized by the proposed recognition method at rank-1, which demonstrate the robustness of the proposed method against inaccurate alignment.

\subsubsection{Joint Optimization Impacts on Recognition Performance}
\begin{figure}[ht]
	\centering
	\includegraphics[width=0.75\linewidth]{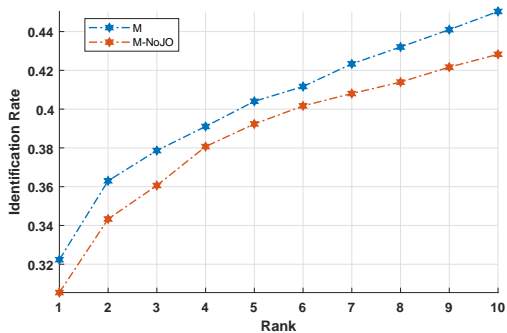}
	\vspace{-0.1in}
	\caption{CMC curves of the methods using the hybrid deep networks trained with and without the joint optimization evaluated on the IIIT-Delhi MOLF database.}
	\vspace{-0.05in}
	\label{fig:optimizationMOLF}
\end{figure}

To verify the effectiveness of the joint optimization, we compare the recognition performance of the hybrid deep network trained with and without the joint optimization evaluated on the IIIT-Delhi MOLF database. Fig. \ref{fig:optimizationMOLF} shows the CMC curves of these two networks trained with and without the joint optimization, which are labeled as M and M-NoJO in the figure, respectively. As can be seen, the performance achieved by the network without the joint optimization is obviously worse than that of the network with the joint optimization, which supports our choice of the joint optimization.

\subsubsection{Visualization of Learned Features}
\label{subsub:LF}
\begin{figure}[htbp]
	\centering
	\includegraphics[width=0.8\linewidth]{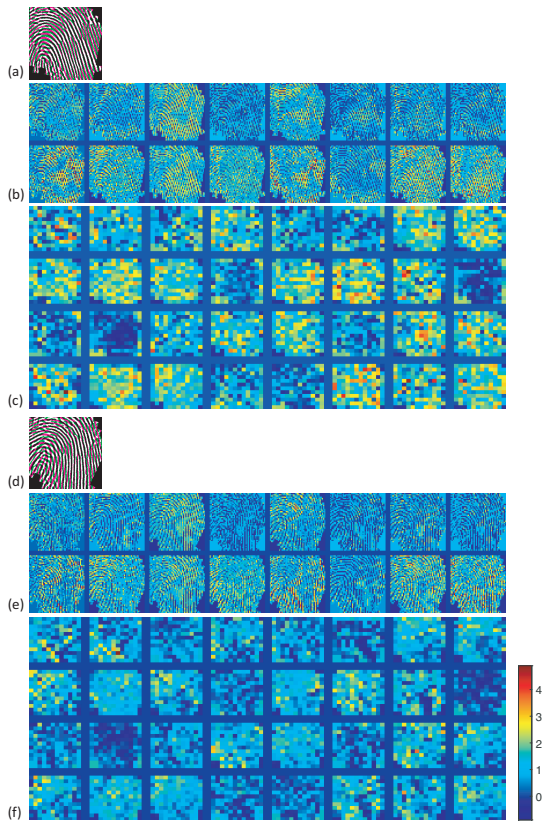}
	\vspace{-0.1in}
	\caption{Visualization of the features learned by the first and second layers of the cCNN$_{4,1}$: (a) a true pair of fingerprints; (b) features of (a) learned by the first layer; (c) features of (a) learned by the second layer; (d) a false pair of fingerprints; (e) features of (d) learned by the first layer; and (f) features of (d) learned by the second layer.}
	%\vspace{-0.1in}
	\label{fig:features}
\end{figure}

A visualization of the features learned by the first and second layers of the cCNN$_{4,1}$ is shown in Fig. \ref{fig:features}. As can be seen, the proposed deep network tends to learn the similarity of two fingerprint images. The higher the similarity, the higher the response in the feature maps. For a genuine pair of fingerprints, most of its feature maps have a high response, and for a false pair of fingerprints, most of its feature maps have a low response. This shows the interpretability and the soundness of the proposed deep network and the learned features.

\vspace{-0.1in}
\subsubsection{Running Times}
\label{subsubsec:time}
The hybrid deep network is trained off-line on the Australia National Computational Infrastructure (NCI). The training time for each component and the overall hybrid deep network is provided in Appendix E. The recognition is run on Intel(R) Xeon(R) E5-2670 CPUs (2.6 GHz) powered by a Linux system. On the recognition stage, using the pre-trained network to perform prediction for each fingerprint pair about 0.2s. This time cost is mainly due to the processing of the large number of patches and macro-patches by the set of CNNs. To reduce the computational burden caused by this, we use 24 threads to process the patches and marco-patches for different CNNs in parallel, which can reduce the running time to less than 0.02s.

\vspace{-0.05in}
\section{Conclusion}
\label{sec:conclusion}
This paper proposed a new scheme for latent fingerprint recognition. Instead of extracting representation features of each fingerprint independently and then compare the similarity of these representation features for recognition in a different process, we proposed to directly model the pair-relationship of two fingerprints as the similarity feature for recognition. This way, correlations of two fingerprints are exploited for better characterization of the similarity which is the base of matching decision making. The pair-relationship is modeled by a hybrid deep network using FIT and OFT tensors of two fingerprints which can provide complementary information. The hybrid deep network is delicately designed to have a set of CNNs with different sizes and architectures to handle the difficulties of random sizes and corrupted areas of latent fingerprints. Experimental results evaluated on two databases demonstrate that the proposed method outperforms the state of the art.

\ifCLASSOPTIONcompsoc
\section*{Acknowledgments}
\else
\section*{Acknowledgment}
\fi
This project is supported by ARC Discovery Grant with project ID DP190103660 and ARC Linkage Grant with project ID LP180100663.

\ifCLASSOPTIONcaptionsoff
\newpage
\fi

\bibliographystyle{IEEEtran}
\bibliography{PAMI_latent_recognition}
\vspace{-0.3in}
\end{document}